\documentclass[runningheads]{llncs}
\usepackage[T1]{fontenc}
\usepackage[utf8]{inputenc}
\usepackage{balance}
\usepackage[table]{xcolor}
\usepackage{amsmath,amssymb,amsfonts}
\usepackage{textcomp}
\usepackage{booktabs}
\usepackage{multirow}
\usepackage{todonotes}
\usepackage{tabularx}
\usepackage{subfig}
\usepackage{graphicx}
\usepackage{marvosym}

\usepackage{algorithmic}
\usepackage{color, colortbl}
\usepackage{sidecap}
\usepackage{rotating}
\usepackage[ruled,vlined,linesnumbered,noresetcount]{algorithm2e}
\SetKwInput{KwInput}{Input}                
\SetKwInput{KwOutput}{Output}              
\newcommand{\croppdf}[1]{\IfFileExists{#1-crop.pdf}{}{\immediate\write18{pdfcrop #1.pdf}}}
\newsavebox\CBox

\usepackage[hypertexnames=false]{hyperref} 
\usepackage[capitalise,nameinlink,noabbrev]{cleveref}
\hypersetup{
 colorlinks=true,
 linkcolor=blue,
 urlcolor=blue,
 citecolor=blue
}
\usepackage{scalefnt}
\usepackage{pgfplots}
\DeclareUnicodeCharacter{2212}{−}
\usepgfplotslibrary{groupplots,dateplot}
\usetikzlibrary{patterns,shapes.arrows}
\pgfplotsset{compat=newest}
\usepackage{lmodern}
\usepackage{changepage}
\usepackage{bbm}
\usepackage{orcidlink} 
\definecolor{LightOrange}{rgb}{1, 0.96, 0.87}
\definecolor{LightCyan}{rgb}{0.88,1,1}
\Crefname{equation}{Eq.}{Eqs.}
\newcolumntype{h}{>{\columncolor{LightOrange}}c}
\newcolumntype{k}{>{\columncolor{LightOrange}}l}

\newcommand{\mcS}{\mathcal{S}}

\newcommand{\subbest}[1]{{\textbf{{#1}}}}

\newcommand{\etal}{\textit{et~al.}}

\usepackage{listings}
\begin{document}
\Crefname{figure}{Suppl. Figure}{Suppl. Figures}
\Crefname{table}{Suppl. Table}{Suppl. Tables}
\Crefname{section}{Suppl. Section}{Suppl. Sections}
\Crefname{proposition}{Suppl. Proposition}{Suppl. Propositions}
\Crefname{algorithm}{Suppl. Algorithm}{Suppl. Algorithms}

\title{SiNGR: Brain Tumor Segmentation via Signed Normalized Geodesic Transform Regression}
\titlerunning{SiNGR: Signed Normalized Geodesic Transform Regression}

\author{Trung DQ. Dang\thanks{Equal contributions}
\and
Huy Hoang Nguyen$^{\star}$ \and
Aleksei Tiulpin}

\authorrunning{Accepted as a conference paper at MICCAI 2024}
\institute{University of Oulu, Finland \\
\email{\{trung.ng,huy.nguyen,aleksei.tiulpin\}@oulu.fi}}

\maketitle

\begin{abstract}
One of the primary challenges in brain tumor segmentation arises from the uncertainty of voxels close to tumor boundaries. However, the conventional process of generating ground truth segmentation masks fails to treat such uncertainties properly. Those ``hard labels'' with 0s and 1s conceptually influenced the majority of prior studies on brain image segmentation. As a result, tumor segmentation is often solved through voxel classification. In this work, we instead view this problem as a voxel-level regression, where the ground truth represents a certainty mapping from any pixel to the border of the tumor. We propose a novel ground truth label transformation, which is based on a signed geodesic transform, capturing the uncertainty of brain tumors' vicinity. We combine this idea with a Focal-like regression L1-loss that enables effective regression learning in high-dimensional output space by appropriately weighting voxels according to their difficulty. We thoroughly conduct an experimental evaluation, validating the components of our proposed method, comparing it to a diverse array of state-of-the-art segmentation models, and showing that it is architecture-agnostic. The code of our method is made publicly available (\url{https://github.com/Oulu-IMEDS/SiNGR/}). 

\keywords{Semantic Segmentation \and Soft Labels \and Brain Tumor \and Signed Geodesic Transform}
\end{abstract}

\section{Introduction}
Glioma is the most prevalent type of brain tumor in adults, which is also the leading cause of cancer deaths for men under 40 years old and women under 20 years old~\cite{siegel2022cancer}. Timely detection and characterization of gliomas is crucial for patient survival. Fortunately, the widespread availability of magnetic resonance imaging (MRI) enables non-invasive quantitative brain assessments, enabling healthcare professionals to detect and closely monitor the progression of brain tumors~\cite{verduin2018noninvasive}. In response to this, various studies have been conducted to develop deep learning (DL)-based methods for segmentation of brain tumors' volumes from  MR images~\cite{vasudeva2024geols,hatamizadeh2022unetr,hatamizadeh2021swin,wang2023dice}. In this paper, so as it is done conventionally, by segmentation we imply semantic segmentation (SSEG), the goal of which is to categorize each element (voxel or pixel) within an image into background and foreground classes. 

In the existing literature on brain tumor segmentation, the first line of research aims to improve the effectiveness of DL architectures by increasing the model capacity and/or embedding domain knowledge into architecture design~\cite{hatamizadeh2021swin,she2023eoformer,xing2022nestedformer,wang2021transbts,hatamizadeh2022unetr}. The second line of research studies novel segmentation losses that are preferably correlated with metrics of interest such as Dice score or Intersection-of-Union (IoU)~\cite{bertels2019optimizing,berman2018lovasz,salehi2017tversky,wang2023dice,wang2023jaccard}. Finally, the third group of studies investigates alternative ways to define the ground truth (GT) masks through \emph{soft labels}~\cite{vasudeva2024geols,ma2023enhanced}. Our study is at the intersection of the last two directions, as we aim to define the GT masks through soft labels, as well as to develop a new loss for the problem. 

SSEG is typically modeled as voxel-wise \emph{classification}~\cite{hatamizadeh2021swin,she2023eoformer,xing2022nestedformer,wang2021transbts,hatamizadeh2022unetr}. This might be inherited from the annotation process of GT masks, where each voxel is assigned to one or more classes by human annotators. However, such strictly defined categories (e.g. $0$ vs $1$ in the binary case) in the segmentation masks\footnote{Hereinafter, we use the terms segmentation masks, 0-1 GT masks, and hard labels interchangeably.} imply an equal role for all voxels when it comes to determining the edges of an object of interest in an image. Such an approach fails to capture the intra-class uncertainty in the annotation masks. Delineating complex non-rigid structures such as tumors in low-resolution MR images is highly challenging, and the uncertainty one can easily identify could come from technical image quality, visibility, detail complexity, and the knowledge of the annotator. 

\begin{figure}[t]
    \centering
    \croppdf{figures/SiNGR_workflow}    
    \includegraphics[width=\textwidth]{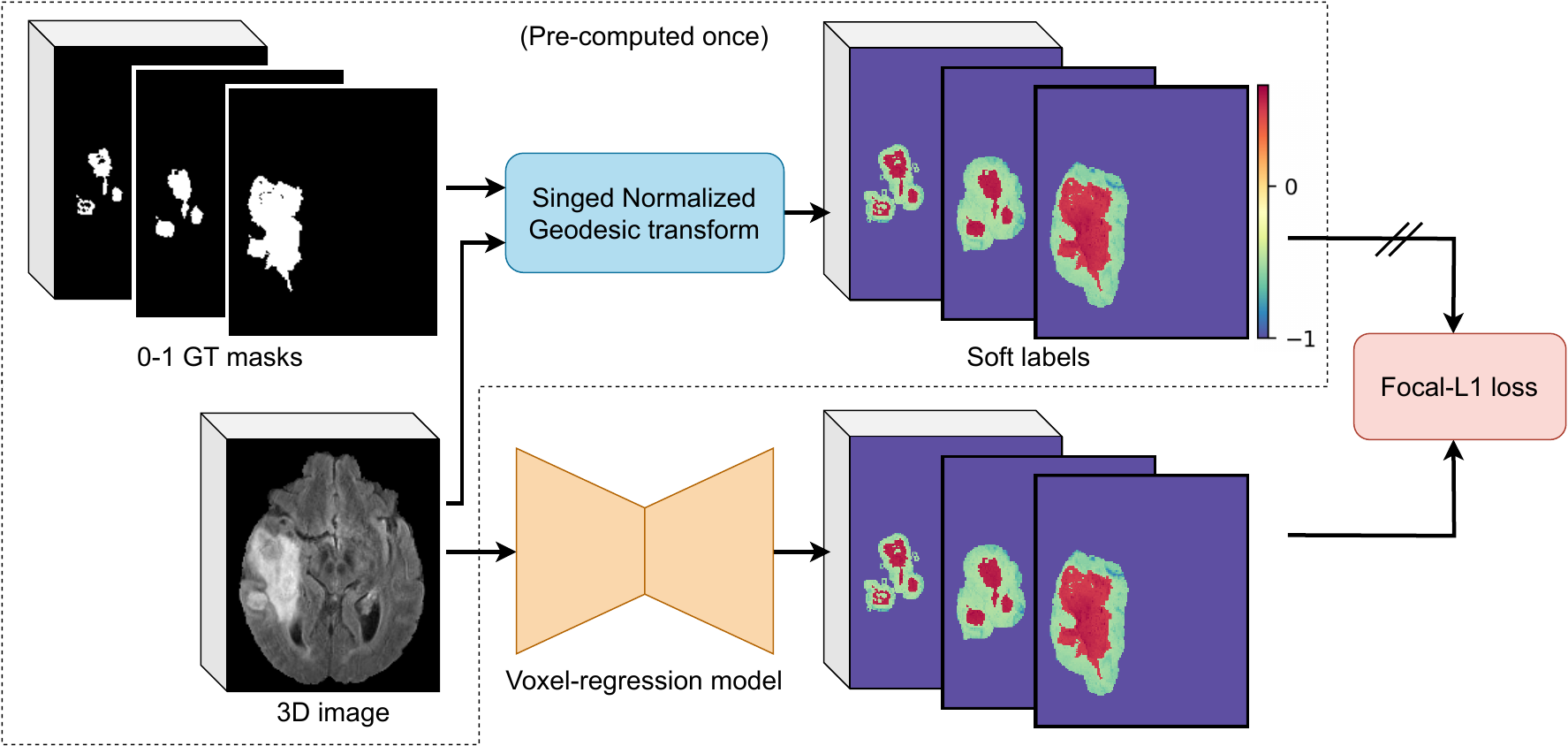}
    \caption{The overview of Signed Normalized Geodesic transform Regression (SiNGR).}
    \label{fig:workflow}
\end{figure}

Some studies~\cite{vasudeva2024geols,ma2023enhanced,wang2023dice,wang2023jaccard} attempted to develop soft labels for SSEG, which allows the model to learn that it should not always be a certainty. One such example is label smoothing.
This technique relies on the number of classes~\cite{szegedy2016rethinking}, which is not suitable for the aforementioned intra-class uncertainty. Liu~\etal~\cite{liu2022combining} proposed to produce soft labels using the signed Euclidean distance map (SDM). Nevertheless, the SDM regression was merely considered as a regularizer for the segmentation task. Vasudeva~\etal~\cite{vasudeva2024geols} employed the unsigned geodesic distance (GeoDT) transform~\cite{toivanen1996new} to generate a novel type of soft labels that can characterize both spatial distance and image gradient. Yet unsigned GeoDT-based soft labels were then integrated into the cross-entropy (CE) loss, which is a classification loss.

We observe that the complexity of the segmentation annotation process is in the labeling of voxels around the boundaries of objects of interest (OOI). The uncertainty of a voxel's label is proportional to its distance to the nearest boundary, as well as the visual blurriness in this region. Although unsigned GeoDT naturally allows us to capture these properties, additional signals are needed to differentiate foreground from background voxels. Moreover, voxels significantly distant from the OOI exhibit very low uncertainty; thus, these voxels should be marked in a manner that directs the model to pay less attention to them.

In this study, we formulate the SSEG problem as \emph{voxel regression} (see~\cref{fig:workflow}). We propose an extension of the unsigned GeoDT~\cite{toivanen1996new}, termed \emph{Signed Normalized Geodesic} (SiNG) transform that aims to approximate modeling of segmentation annotations done by human from input images and the corresponding GT masks. The SiNG transform is designed to primarily focus on the vicinity of the OOI by assigning values in $(0, 1]$ to foreground (FG) regions, values in $(-1, 0]$ to nearby background (BG) regions, and $-1$'s to distant voxels. As we perform SSEG via \emph{SiNG transform Regression}, our method is named SiNGR. To handle the imbalance of FG and BG voxels, we introduce a novel regression loss, termed Focal-L1 loss. We conduct standardized and thorough experiments to demonstrate the effectiveness of our method on the BraTS and LGG FLAIR datasets. The empirical evidence shows that our method is beneficial for various DL architectures.

\section{Method}
\subsection{Overview}
We approach the SSEG problem via signed soft-label regression. Specifically, we utilize the SiNG transform presented in~\cref{sec:sing} to convert 0-1 GT masks to signed soft labels, where regions of interest and BG regions are represented by positive and negative values respectively. The proposed transform is designed in such a way that FG and BG voxels are marginally separate across $0$, which allows a simple $0$-threshold post-processing step to produce final predicted masks.  To effectively train the regression task, we introduce a novel loss, named \emph{Focal-L1}, in~\cref{sec:focall1_loss}.

\subsection{Signed Normalized Geodesic Transform}
\label{sec:sing}
\subsubsection{Unsigned transform.}
Given an input image $I$ with  spatial dimensions of $H \times W \times L$, and an arbitrary-shaped region $R \subset \Omega=[H]\times[W]\times[L]$ with $[N]=\{1, \dots, N\}$, an L1-based \emph{unsigned geodesic distance transform} (GeoDT) from a point $i \in \Omega$ to $R$ is defined as~\cite{toivanen1996new,wang2018deepigeos}
\begin{align}
    G^\lambda(i; R, I) &= \min_{j \in R} D^\lambda(i,j, I), \label{eq:geodt_1} \\
    D^\lambda(i,j; I) &= \min_{p \in P_{i,j}} \int_{0}^{1} (1-\lambda)\|p'(s) \|_1 + \lambda \| \nabla I(p(s)) \cdot u(s) \|_1 \label{eq:geodt_2}
\end{align}
where $\lambda \in [0,1]$ is a weighting hyperparameter, $P(i,j)$ is the set of all paths between locations $i$ and $j$, $p$ is a feasible parameterized path, $u(s)=\frac{p'(s)}{\| p'(s)\|_1}$, and $\nabla I(p(s))$ represents image gradient at $p(s)$. Here, we have that $G^\lambda(i, R, I) = 0 \ \forall i \in R, \ \forall \lambda \in [0,1]$. Intuitively, the unsigned GeoDT calculates the cost of the shortest path from point $i$ to the region $R$ based on both distance and image gradient information. The integral makes GeoDT computationally expensive when we apply the transform for the whole image. Its cost is proportional to $R$'s cardinality for~\cref{eq:geodt_1} and $I$'s size due to~\cref{eq:geodt_2}. 

\subsubsection{Signed version.}
\label{sc:sgeo_uncertainty}
For \emph{signed GeoDT}, one typically runs the GeoDT transform twice for the foreground and background regions, which is intensively costly~\cite{fu2016geodesic}. To speed up the signed GeoDT, we merely consider a set $R_{B}$ of boundary voxels of OOI extracted from a 0-1 segmentation mask $M$ using the Canny edge detector~\cite{canny1986computational}. Hereinafter, as $R_{B}$ is our primary region of interest, we omit $R_{B}$ and $I$ from~\cref{eq:geodt_1} for simplicity, that is $G^\lambda_i=G^\lambda(i; R_{B}, I)$. Given $R_{B}$, we apply unsigned GeoDT for the whole image to produce an unsigned map. Afterwards, we rely on $M$ to specify the signs of the obtained map, that is $s_i G^\lambda_i, \forall i \in \Omega$, where $s_i = \mathrm{sign}(2M_i - 1)$ is the sign of voxel $i$.

As the uncertainty of human annotations primarily is around the boundary $R_{B}$, we thus ignore regions substantially far from the boundary. As such, we let $\lambda = 0$, and define the neighboring region of the boundary 
\begin{align}
    \mathcal{B} &= \left \{ i \in \Omega \mid G^0_i \leq \beta \right \} \\
    \mathrm{s.t.} \ \beta &= \max_{i \in \Omega: M_i=1} G^0_i    
\end{align}
where $\beta$ represents the maximum spatial distance (i.e.\ the latter term in~\cref{eq:geodt_2} is omitted when $\lambda = 0$). Finally, we propose the SiNG transform as follows
\begin{align}
   \mcS_i = \left\{\begin{matrix}
 \frac{1}{\tau} (1- \delta) s_i G^\lambda_i  + \delta s_i & \quad \ \mathrm{if} \ i \in \mathcal{B} \\ 
 -1 & \quad \ \mathrm{otherwise}
 \end{matrix}\right. 
\label{eq:sng}
\end{align}
where $\tau = \max_{j \in \mathcal{B}} G^\lambda_j$ is a normalization constant for each pair $(I, M)$, and $\delta\in[0, 1)$ is a margin hyperparameter. The margin $\delta$ is needed to ensure the discrepancy in the signed normalized map between foreground and background pixels.

\begin{figure}[t]
    \centering
    \croppdf{figures/l1_loss}
    \croppdf{figures/focall1_loss}
    \croppdf{figures/focall1_gamma}    
    \subfloat[L1 loss]{
    \includegraphics[width=0.32\textwidth]{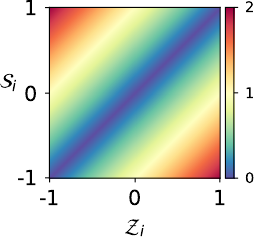}}
    \hfill
    \subfloat[Focal-L1 loss ($\gamma=1$)]{
    \includegraphics[width=0.32\textwidth]{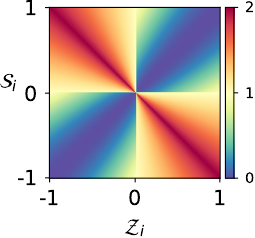}}
    \hfill
    \subfloat[Focal-L1 behavior]{
    \includegraphics[width=0.33\textwidth]{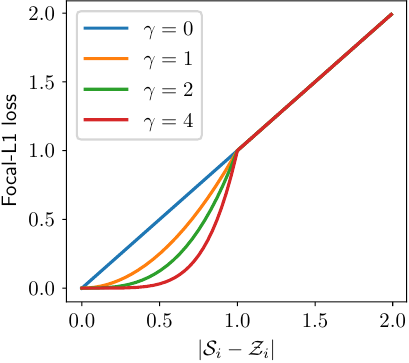}}
    \caption{Comparison between L1 and Focal-L1 losses: (a-b) 2D loss surfaces, and (c) a behavior of the proposed Focal-L1 loss. Colors in (a-b) represent loss magnitudes. In (c), we assume that $|\mcS_i| = 1$ or $|\mathcal{Z}_i|=1, \forall i \in \Omega$.    
    }
    \label{fig:focalf1_2d_surface}
\end{figure}

\subsection{Focal-L1 Loss}
\label{sec:focall1_loss}
We formulate the tumor segmentation task as image-level ``regression'' rather than voxel ``classification'', where one commonly utilizes the cross-entropy (CE) or focal loss for the optimization. In alignment with the preceding subsection, we also consider the significance of diversity across various regions. As such, the regression loss is supposed to prioritize hard regions over easy ones. Inspired by the CE-based focal loss~\cite{lin2017focal}, we propose the following L1-based focal loss, namely Focal-L1, for the regression task.

Given an arbitrary pair of inputs $(I, M)$, we utilize the SiNG transformation to produce a corresponding map $\mcS$ with a certain $\lambda$.
In addition, assuming that we have a parametric function $f_\theta$ with parameters $\theta$, and $f_\theta(I)$ denotes a predicted mask from the input image $I$. We then utilize the tanh function to convert values of $f_\theta(I)$ into the range $[-1, 1]$, that is $\mathcal{Z}=\tanh(f_\theta(I))$. 
To this end, we propose the Focal-L1 loss as follows
\begin{equation}
    \mathcal{L}_{\mathrm{FocalL1}} (\mcS, \mathcal{Z}; \theta) = \frac{1}{| \Omega |} \sum_{i \in \Omega} |\mcS_i - \mathcal{Z}_i| \underbrace{\frac{|\mcS_i - \mathcal{Z}_i|^{\gamma \mathbbm{I}(\mcS_i \mathcal{Z}_i \geq 0)}}{\max(|\mcS_i|, |\mathcal{Z}_i|) + \varepsilon} }_{\mathrm{Sample\ weighting}}, \label{eq:focal_l1_loss}
\end{equation}
where $\varepsilon$ is a positive constant to avoid numerical issues, $\gamma$ is a positive hyperparameter, and $\mathbb{I}(\cdot)$ is the indicator function. A graphical comparison between our loss and the L1 loss is presented in~\cref{fig:focalf1_2d_surface}. 
Note that the backpropagation is not applied to the weighting term. Whereas the weighting term's numerator is fixed at $1$ for hard cases ($\mcS_i\mathcal{Z}_i < 0$), it is scaled down to less than $1$ for easy ones ($\mcS_i\mathcal{Z}_i \geq 0$). Meanwhile, the denominator reduces the importance of highly certain voxels  (with high absolute value), as they are 
straightforward to predict. Overall, the sample weighting helps the loss to prioritize enforcing hard voxels over simple ones.

\section{Experiments}

\begin{table}[t]
    \centering    
    \renewcommand{\arraystretch}{1.2}
    \caption{Performance comparisons between our methods and the SOTA baselines on the BraTS test set (means and SEs over $5$ random seeds). The best results with substantial differences are highlighted in bold.
    }
\resizebox{\textwidth}{!}{
\begin{tabular}{|l|c|c|c|c|c|c|c|c|c|}
\hline 
      \multirow{2}{*}{Method} &      \multirow{2}{*}{GT} &        \multicolumn{4}{c|}{Dice score (\%) $\uparrow$} &             \multicolumn{4}{c|}{HD95 ($mm$) $\downarrow$} \\
      \cline{3-10} 
       &         &     ET &              TC &              WT &             Avg &             ET &             TC &             WT &            Avg \\
\hline  
    TransBTS~\cite{wang2021transbts} & \multirow{8}{*}{\rotatebox[origin=c]{90}{Hard label}} & 81.0$_{\pm0.3}$ & 83.0$_{\pm0.4}$ & 90.4$_{\pm0.1}$ & 84.8$_{\pm0.2}$ & 4.2$_{\pm0.3}$ & 6.4$_{\pm0.2}$ & 6.4$_{\pm0.1}$ & 5.7$_{\pm0.2}$ \\
    SegResnet~\cite{myronenko20193d} & & 81.1$_{\pm0.3}$ & 85.5$_{\pm0.3}$ & 90.7$_{\pm0.1}$ & 85.8$_{\pm0.2}$ & 3.3$_{\pm0.1}$ & 5.5$_{\pm0.3}$ & 5.7$_{\pm0.4}$ & 4.9$_{\pm0.2}$ \\
    UNETR~\cite{hatamizadeh2022unetr} & & 82.3$_{\pm0.2}$ & 82.0$_{\pm0.4}$ & 90.0$_{\pm0.1}$ & 84.8$_{\pm0.2}$ & 4.0$_{\pm0.3}$ & 7.4$_{\pm0.2}$ & 6.2$_{\pm0.4}$ & 5.9$_{\pm0.2}$ \\
    EoFormer~\cite{she2023eoformer} & & 82.5$_{\pm0.2}$ & 84.8$_{\pm0.4}$ & \subbest{91.3$_{\pm0.0}$} & 86.2$_{\pm0.2}$ & 3.7$_{\pm0.2}$ & 6.4$_{\pm0.4}$ & 5.9$_{\pm0.3}$ & 5.3$_{\pm0.2}$ \\
    UNet++3D~\cite{zhou2019unet++} & & 83.1$_{\pm0.1}$ & 86.0$_{\pm0.2}$ & 90.9$_{\pm0.2}$ & 86.7$_{\pm0.1}$ & 4.4$_{\pm0.4}$ & 6.3$_{\pm0.3}$ & 5.7$_{\pm0.3}$ & 5.5$_{\pm0.3}$ \\
    UNet3D~\cite{kerfoot2019left} & & 83.1$_{\pm0.2}$ & 86.1$_{\pm0.3}$ & 90.4$_{\pm0.2}$ & 86.5$_{\pm0.1}$ & 3.8$_{\pm0.4}$ & 5.9$_{\pm0.3}$ & 6.5$_{\pm0.6}$ & 5.4$_{\pm0.4}$ \\  
    NestedFormer~\cite{xing2022nestedformer}&  & 83.5$_{\pm0.1}$ & 85.4$_{\pm0.1}$ & \subbest{91.2$_{\pm0.1}$} & 86.7$_{\pm0.1}$ & 4.4$_{\pm0.4}$ & 7.4$_{\pm0.4}$ & 6.6$_{\pm0.4}$ & 6.1$_{\pm0.4}$ \\
    Swin-UNETR~\cite{hatamizadeh2021swin} & & 84.1$_{\pm0.2}$ & 85.7$_{\pm0.4}$ & 90.9$_{\pm0.1}$ & 86.9$_{\pm0.2}$ & 3.7$_{\pm0.3}$ & 6.4$_{\pm0.2}$ & 6.0$_{\pm0.2}$ & 5.4$_{\pm0.2}$ \\
    \hline
    Ours (Swin-UNETR) & SiNG & \subbest{85.1$_{\pm0.3}$} & \subbest{88.0$_{\pm0.1}$} & \subbest{91.3$_{\pm0.1}$} & \subbest{88.1$_{\pm0.1}$} & \subbest{2.3$_{\pm0.1}$} & \subbest{4.2$_{\pm0.2}$} & \subbest{4.8$_{\pm0.0}$} &\subbest{3.8$_{\pm0.1}$} \\
\hline
\end{tabular}    
}
    \label{tab:brats_results}
\end{table}

\subsection{Setup}

\noindent\textbf{Datasets.} We thoroughly conducted experiments on two public brain tumor datasets: BraTS 2020 and LGG FLAIR. \textbf{BraTS 2020}~\cite{menze2014multimodal} consists of multi-modal MR images from $369$ subjects. Each image was aligned across four modalities and was standardized to a volume size of $240\times240\times155$. The segmentation targets are enhancing tumor (ET), tumor core (TC), and whole tumor (WT). We divided the dataset into three portions with $236$, $59$, and $74$ samples for training, validation, and test sets, respectively. \textbf{LGG FLAIR}~\cite{buda2019association} has $110$ 3-channel FLAIR MR images and corresponding abnormality segmentation masks. The number of axial slices of each MR scan varies from $20$ to $80$, and they have a common spatial dimension of $256\times256$. We split the data into training, validation, and test sets with $70$, $18$, and $22$ samples, respectively.

\noindent\textbf{Implementation details.} All models were trained on Nvidia V100 GPUs. We implemented our pipeline using Pytorch and followed a standard data processing configuration for all methods. We used the FastGeodist library to compute unsigned GeoDT maps~\cite{asad2022fastgeodis}. During the training process on BraTS, the 3D MR images were randomly cropped to $128\times128\times128$ cubes, and augmented by flipping, intensity scaling, and intensity shifting. We performed the window-slicing technique with a window size of $128\times128\times128$ in the validation and test stages. The employed patch size on LGG was $128\times128\times32$. We utilized the Adam optimizer with an initial learning rate of $10^{-4}$, a weight decay of $10^{-5}$, and a batch size of $2$. Regarding SiNGR-specific hyperparameters, we empirically found $\lambda = 0.5$, $\delta = 0.5$, and $\gamma = 1$ working the best. The hyperparameter $\varepsilon$ in~\cref{eq:focal_l1_loss} was set to $0.0$. We performed $0$-thresholding to binarize our predicted maps. For performance evaluation, we utilized image-wise IoU, Dice score, and $95\%$ Hausdorff distance (HD95).

We compared our method to a diverse array of references specializing in 3D data as listed in~\cref{tab:brats_results,tab:lgg_results}. In addition, we validated the SiNG transform and the Focal-L1 loss against other soft-label baselines -- namely label smoothing (LS)~\cite{szegedy2016rethinking} and Geodesic label smoothing (GeoLS)~\cite{vasudeva2024geols} -- together with the Jaccard Metric Loss (JML)~\cite{wang2023jaccard}, which specializes in optimizing with soft labels.

\begin{table}[t]
    \centering
    \caption{Performance comparisons between our methods and the SOTA baselines on the LGG FLAIR test set (means and SEs over $5$ random seeds). The best results are highlighted in bold.}
    \setlength{\tabcolsep}{6.4pt}
    \renewcommand{\arraystretch}{1.2}
    \begin{tabular}{|l|c|c|c|c|}
\hline
      Method &            GT &    IoU (\%) $\uparrow$ &              Dice (\%) $\uparrow$ &              HD95  ($mm$) $\downarrow$               \\
\hline
EoFormer~\cite{she2023eoformer} & \multirow{8}{*}{\rotatebox[origin=c]{90}{Hard label}} & 38.8$_{\pm11.3}$ & 68.4$_{\pm3.0}$ & 27.1$_{\pm5.1}$  \\
NestedFormer~\cite{xing2022nestedformer} & & 46.6$_{\pm3.1}$ & 60.4$_{\pm1.1}$ &  58.4$_{\pm1.7}$  \\   
UNETR~\cite{hatamizadeh2022unetr} & & 49.5$_{\pm3.2}$ & 62.8$_{\pm1.0}$ &  49.5$_{\pm1.8}$ \\
SegResnet~\cite{myronenko20193d} & & 50.5$_{\pm1.5}$ & 64.9$_{\pm0.5}$ & 46.3$_{\pm2.1}$  \\      
UNet++3D~\cite{zhou2019unet++} & & 59.0$_{\pm3.3}$ & 71.1$_{\pm1.1}$ &  34.6$_{\pm1.3}$  \\
Swin-UNETR~\cite{hatamizadeh2021swin} & & 54.4$_{\pm2.3}$ & 67.8$_{\pm0.8}$ & 32.2$_{\pm1.8}$ \\
TransBTS~\cite{wang2021transbts} & & 58.5$_{\pm1.1}$ & 71.9$_{\pm0.5}$ & 24.8$_{\pm2.4}$ \\    
UNet3D~\cite{kerfoot2019left} & & 57.9$_{\pm4.0}$ & 69.1$_{\pm1.3}$ &  33.6$_{\pm2.0}$  \\    
  \hline      
  Ours (UNet3D) & SiNG & \subbest{71.4$_{\pm0.7}$} & \subbest{82.3$_{\pm0.5}$} & \subbest{11.4$_{\pm1.6}$} \\
\hline
\end{tabular}
    \label{tab:lgg_results}
\end{table}

\begin{figure}[t]
    \centering
    \croppdf{figures/SiNGR_improvements_BRATS_Dice}
    \croppdf{figures/SiNGR_improvements_BRATS_HD95}
    \croppdf{figures/SiNGR_improvements_LGG_Dice}
    \croppdf{figures/SiNGR_improvements_LGG_IoU}
    \croppdf{figures/SiNGR_improvements_LGG_HD95}
    \subfloat[Dice on BraTS\label{fig:impact_singr_on_brats_1}]{\includegraphics[height=0.24\textwidth]{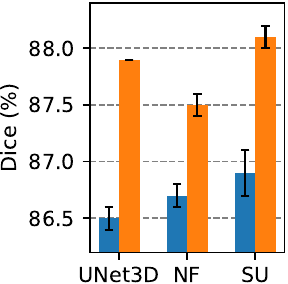}}
    \hfill
    \subfloat[HD95 on BraTS\label{fig:impact_singr_on_brats_2}]{\includegraphics[height=0.24\textwidth]{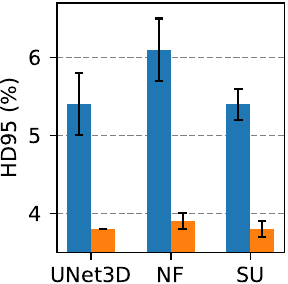}}
    \hfill
    \subfloat[Dice on LGG\label{fig:impact_singr_on_lgg_1}]{\includegraphics[height=0.24\textwidth]{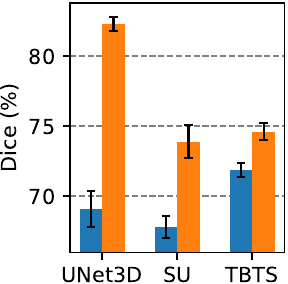}}
    \hfill
    \subfloat[HD95 on LGG\label{fig:impact_singr_on_lgg_2}]{\includegraphics[height=0.24\textwidth]{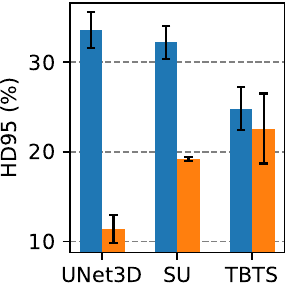}}
    \caption{Pair-wise comparisons between baselines using hard labels (in blue) and our method (in orange) across different architectures and metrics. NF, SU, and TBTS mean NestedFormer, Swin-UNETR, and TransBTS respectively.}
    \label{fig:impact_singr_on_arch}
\end{figure}

\subsection{Results}

\noindent\textbf{BraTS 2020 dataset.} The quantitative results on the BraTS test set are presented in~\cref{tab:brats_results}. Accordingly, the top-3 methods with the highest IoU used the UNet3D, NestedFormer, and Swin-UNETR architectures. When we applied our SiNGR method to those three architectures, we observed consistent and substantial improvements in all the metrics (see~\cref{fig:impact_singr_on_brats_1,fig:impact_singr_on_brats_2}). For instance, compared to the best baseline, Swin-UNETR, our corresponding method achieved $1.2\%$ and $1.6mm$ better in average Dice score and average HD95, respectively. We present a qualitative comparison of the methods in \textcolor{blue}{Suppl. Figure S1}.

\noindent\textbf{LGG FLAIR dataset.}
We present the detailed results on the LGG FLAIR test set in~\cref{tab:lgg_results}. On this dataset, the impact of SiNGR was particularly significant on UNet3D, and this also acquired the highest IoU and Dice score. Compared to the UNet3D reference, our method resulted in significant gains of $13.5\%$ IoU, $13.2\%$ Dice score, and $22.2mm$ HD95. Using our method also consistently led to substantial improvements for Swin-UNETR and TransBTS (see~\cref{fig:impact_singr_on_lgg_1,fig:impact_singr_on_lgg_2}). The qualitative results are presented in \textcolor{blue}{Suppl. Figure S2}.

\noindent\textbf{Impact of SiNG and Focal-L1.}
We investigated the effects of the components of our method and demonstrated the results on the BraTS test set in \cref{tab:ablation_results}. Two unsigned label smoothing techniques~\cite{szegedy2016rethinking,vasudeva2024geols} did not perform well on the task. Compared to LS with JML, our method outperformed with differences of $8.4\%$ and $7.8mm$ in average Dice score and average HD95, respectively. Additionally, among L1, L2, and product~\cite{xue2020shape} losses, L1 was the best combination with the SiNG transform. However, that setting achieved an average Dice of $1.1\%$ lower than our method. Moreover, the empirical evidence showed the importance of the margin in the SiNG transform as well as the sample weighting coefficient in Focal-L1 loss. Notably, excluding the margin $\delta$ led to a substantial drop of $3.4\%$ average Dice score and an increase of $1.5mm$ average HD95.

\begin{table}[t]
    \centering    
    \renewcommand{\arraystretch}{1.2}
    \caption{Performance comparisons between different combinations of soft labels and loss functions on BraTS. UNet3D was the common architecture. $\delta$ is the margin in SiNG. Focal-L1$^{\dag}$ indicates that the sample weighting in~\cref{eq:focal_l1_loss} is simplified to $|\mcS_i - \mathcal{Z}_i|^\gamma$. Our optimal setting is highlighted in cyan.}
    \resizebox{\textwidth}{!}{
\begin{tabular}{|l|c|c|c|c|c|c|c|c|c|c|}
\hline
      \multirow{2}{*}{Label} & \multirow{2}{*}{$\delta$} &\multirow{2}{*}{Loss} & \multicolumn{4}{c|}{Dice score (\%) $\uparrow$} &             \multicolumn{4}{c|}{HD95 ($mm$) $\downarrow$} \\
      \cline{4-11}
       &       &  &     ET &              TC &              WT &             Avg &             ET &             TC &             WT &            Avg \\
\hline 
    LS~\cite{szegedy2016rethinking} & - & JML~\cite{wang2023jaccard} & 75.5$_{\pm1.2}$ & 75.5$_{\pm0.7}$ & 87.5$_{\pm0.3}$ & 79.5$_{\pm0.6}$ & 5.7$_{\pm0.4}$ & 17.8$_{\pm4.5}$ & 11.2$_{\pm1.5}$ & 11.6$_{\pm2.1}$ \\
    GeoLS~\cite{vasudeva2024geols} &  - & JML~\cite{wang2023jaccard} & 67.0$_{\pm0.9}$ & 63.3$_{\pm2.3}$ & 80.8$_{\pm0.7}$ & 70.4$_{\pm1.1}$ & 20.5$_{\pm1.7}$ & 12.2$_{\pm0.9}$ & 24.3$_{\pm2.7}$ & 19.0$_{\pm1.5}$ \\
    \hline    
    \multirow{6}{*}{SiNG} & \multirow{4}{*}{0.5} & L2 & 80.9$_{\pm0.1}$ & 84.3$_{\pm0.2}$ & 90.4$_{\pm0.1}$ & 85.2$_{\pm0.1}$ & 3.9$_{\pm0.4}$ & 5.9$_{\pm0.3}$ & 6.3$_{\pm0.4}$ & 5.4$_{\pm0.3}$ \\ 
    & & Product~\cite{xue2020shape} & 82.5$_{\pm0.4}$ & 86.9$_{\pm0.1}$ & 90.9$_{\pm0.1}$ & 86.8$_{\pm0.2}$ & 3.9$_{\pm0.5}$ & 5.6$_{\pm0.4}$ & 5.5$_{\pm0.3}$ & 5.0$_{\pm0.4}$ \\
    & & L1 & 82.7$_{\pm0.1}$ & 87.1$_{\pm0.3}$ & 90.7$_{\pm0.1}$ & 86.8$_{\pm0.1}$ & 3.3$_{\pm0.3}$ & 4.9$_{\pm0.2}$ & 5.6$_{\pm0.2}$ & 4.6$_{\pm0.2}$ \\   
    & & Focal-L1$^{\dag}$ & 83.4$_{\pm0.3}$ & 87.1$_{\pm0.2}$ & 91.0$_{\pm0.1}$ & 87.1$_{\pm0.2}$ & 3.1$_{\pm0.4}$ & 4.8$_{\pm0.4}$ & 5.2$_{\pm0.2}$ & 4.3$_{\pm0.3}$ \\
    \cline{2-11}   
     & 0 &  & 80.0$_{\pm0.5}$ & 83.8$_{\pm0.5}$ & 89.9$_{\pm0.1}$ & 84.5$_{\pm0.3}$ & 4.3$_{\pm0.3}$ & 5.8$_{\pm0.4}$ & 5.7$_{\pm0.1}$ & 5.3$_{\pm0.3}$ \\
    \rowcolor{LightCyan}
   & 0.5 & \multirow{-2}{*}{Focal-L1} & \subbest{84.4$_{\pm0.1}$} &  \subbest{88.1$_{\pm0.1}$} &  \subbest{91.1$_{\pm0.1}$} &  \subbest{87.9$_{\pm0.0}$} &  \subbest{2.5$_{\pm0.1}$} &  \subbest{4.2$_{\pm0.1}$} & \subbest{4.8$_{\pm0.0}$} & \subbest{3.8$_{\pm0.0}$} \\
\hline
\end{tabular}    
}
    \label{tab:ablation_results}
\end{table}

\section{Conclusion}

We have introduced a simple approach to segmentation of brain tumors through voxel-wise regression. We proposed the novel SiNG transform that allows us to convert 0-1 annotated masks to soft labels that take into account the uncertainty of the labeling process. In addition, we have introduced the Focal-L1 loss to effectively weight voxels according to their difficulty. Our empirical findings indicate that our method consistently enhances performance across different DL architectures.



%
%
%
%
\bibliographystyle{splncs04}
\bibliography{Paper-2261}




\begin{sidewaysfigure}[t] 
    \centering
    \croppdf{figures/viz/plot_BraTS20_Training_011}
    \croppdf{figures/viz/plot_BraTS20_Training_100}
    \croppdf{figures/viz/plot_BraTS20_Training_145}
    \croppdf{figures/viz/plot_BraTS20_Training_187}
    \croppdf{figures/viz/plot_BraTS20_Training_342}
    \croppdf{figures/viz/plot_BraTS20_Training_368}
    \includegraphics[width=\textwidth]{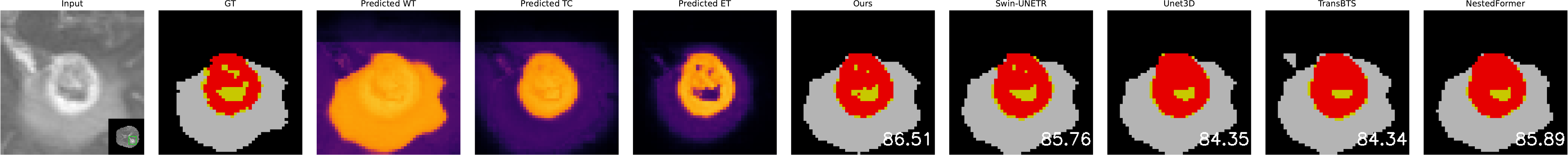} \\
    \includegraphics[width=\textwidth]{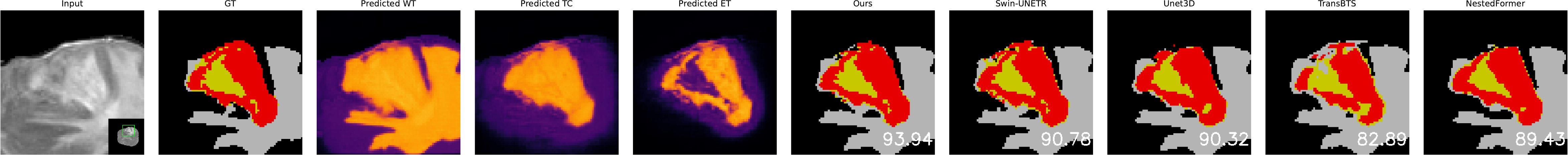} \\
    \includegraphics[width=\textwidth]{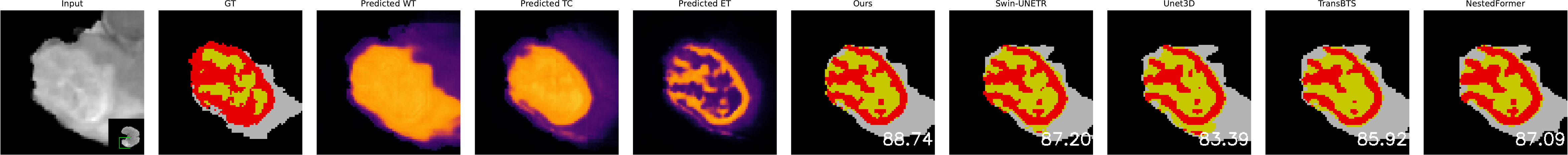} \\
    \includegraphics[width=\textwidth]{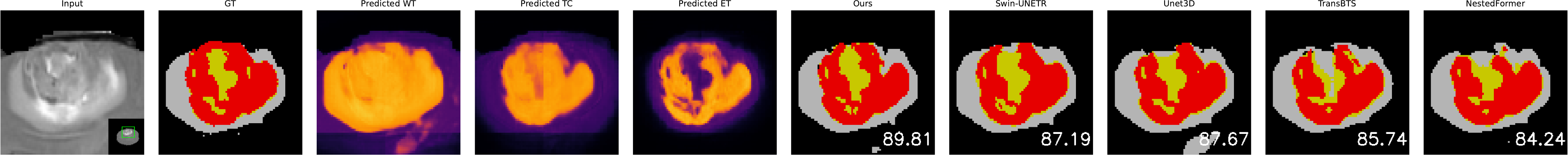} \\
    \includegraphics[width=\textwidth]{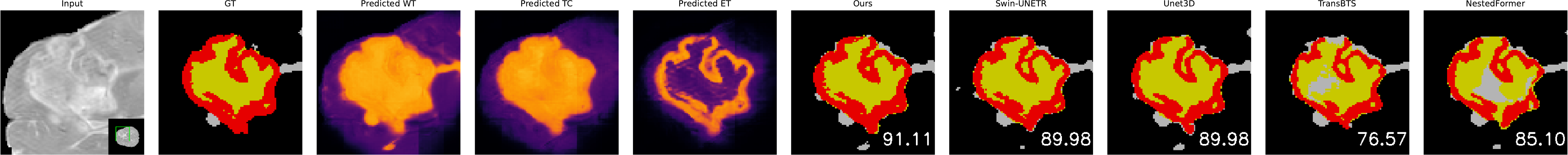} \\ 
    \includegraphics[width=\textwidth]{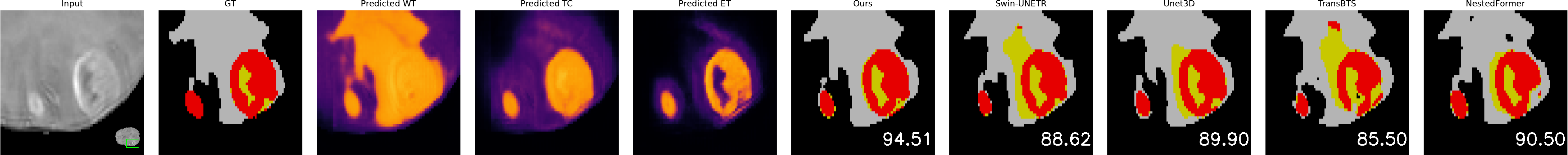}  
    \caption{Visualization of predictions of our method and the baselines on the samples from the BraTS test set. The overlaid numbers are averaged Dice scores over the 3D sample.}
    \label{fig:predictions_viz}
\end{sidewaysfigure}

\begin{sidewaysfigure}[t]
    \centering
    \croppdf{figures/viz_lgg/plot_TCGA_CS_5393_19990606}
    \croppdf{figures/viz_lgg/plot_TCGA_DU_5853_19950823}
    \croppdf{figures/viz_lgg/plot_TCGA_FG_5964_20010511}
    \croppdf{figures/viz_lgg/plot_TCGA_FG_6692_20020606}
    \includegraphics[width=\textwidth]{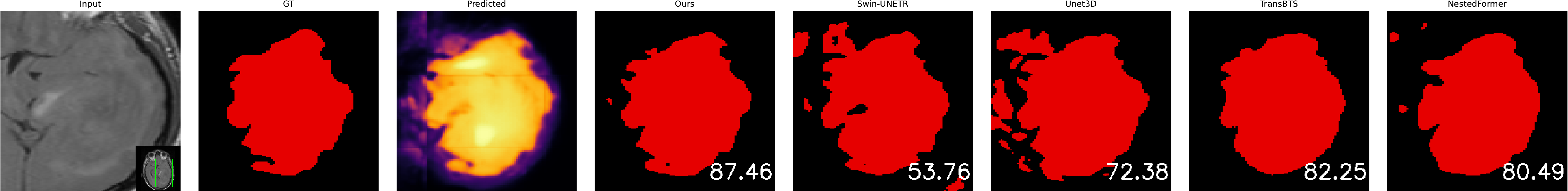} \\
    \includegraphics[width=\textwidth]{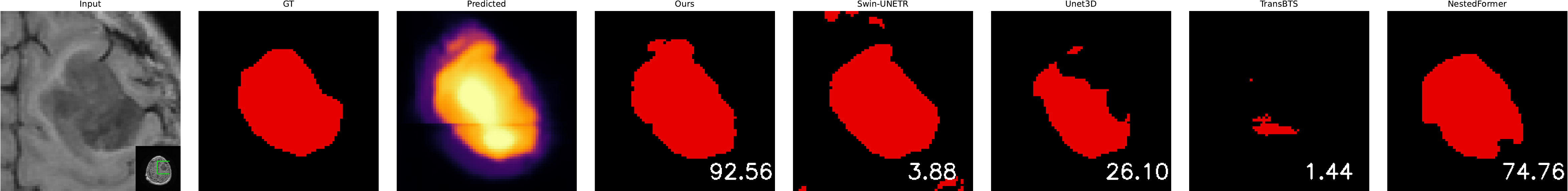} \\
    \includegraphics[width=\textwidth]{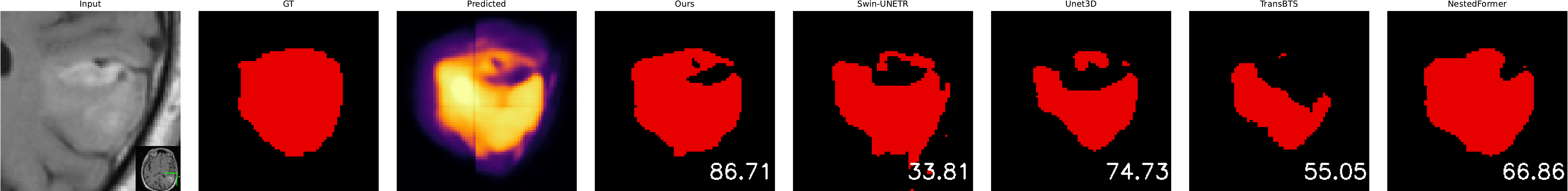} \\
    \includegraphics[width=\textwidth]{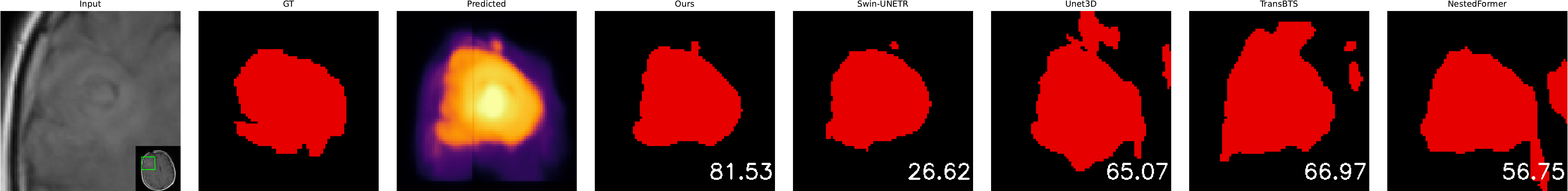}   
    \caption{Visualization of predictions of our method and the baselines on the samples from the LGG FLAIR test set over the 3D sample. }
    \label{fig:predictions_viz_lgg}
\end{sidewaysfigure}

\end{document}